\input amstex
\input xy
\xyoption{all}
\input miniltx
\input graphicx
\input epsf
\documentstyle{amsppt}
\document
\magnification=1200
\NoBlackBoxes
\nologo
\hoffset1.5cm
\voffset2cm
\pageheight {16cm}

\def\cU{\Cal{U}}

\def\R{\bold{R}}
\def\cM{\Cal{M}}
\def\cS{\Cal{S}}
\def\cC{\Cal{C}}
\def\P{\bold{P}}
\def\cN{\Cal{N}}

\bigskip

\centerline{\bf SEMANTIC SPACES}

 \bigskip

\centerline{ Yuri I. Manin${}^1$, Matilde Marcolli${}^2$}

\medskip

\centerline{\it ${}^1$Max--Planck--Institut f\"ur Mathematik, Bonn, Germany,}
\smallskip
\centerline{\it ${}^2$California Institute of Technology, Pasadena, USA}
\vskip1cm

 ABSTRACT. Any natural language can be considered as a tool for
 producing  large databases (consisting of texts, written, or discursive).
 This tool for its description  in turn requires other 
 large databases (dictionaries, grammars etc.). 
 Nowadays, the notion of database is associated with computer processing
 and computer memory. However, a natural language resides also in human brains
 and functions in human communication, from interpersonal to intergenerational one.
 We discuss in this survey/research paper mathematical, in particular geometric,
 constructions, which help to bridge these two worlds. 
 In particular, in this paper we consider the Vector Space Model of semantics 
 based on frequency matrices, as used in Natural Language Processing. We investigate 
 underlying geometries, formulated in terms of Grassmannians, projective spaces, and 
 flag varieties. We formulate the relation between vector space models and semantic 
 spaces based on semic axes in terms of projectability of subvarieties in Grassmannians 
 and projective spaces. We interpret Latent Semantics as a geometric flow on 
 Grassmannians. We also discuss how to formulate G\"ardenfors' notion 
 of  ``meeting of minds" in our geometric setting.

\bigskip
\hfill{ \it O INTERIOR DO EXTERIOR DO INTERIOR }
\smallskip
\hfill{Pascal Mercier}

\hfill{\it ``Nachtzug nach Lissabon''}

\bigskip

\centerline{\bf 1. Introduction: }
\smallskip
\centerline{\bf linguistics, semiotics, and topology}

\bigskip

One of the basic ``meta--physical'' principles of  classical
physics consisted in the subdivision of informational
content of any physical model into two parts:

-- a description of the configuration and phase spaces of
the studied system;

-- a description of the time evolution law (usually a vector field
in the phase space).

\smallskip

Some of the recent approaches to semantics of natural languages
describe various versions of  ``spaces of meanings''  which we consider
as a metaphorical analog of configuration spaces: cf. comprehensive accounts
[G\"a00], [G\"a14]. For G\"ardenfors, semantics is (in particular) {\it meeting of minds,}
and the space of meanings is the space where minds meet.
\smallskip

Our initial motivation for undertaking this survey and the research summarised in this paper
was our desire to introduce  ``a time dimension'' in this
discussion, to see a discourse or reception of a text as a path
in the appropriate space of meanings.

\smallskip

In particular, we wanted to use mathematical models in order 
  to bridge the approaches to semantics reviewed in [G\"a14],
neurolinguistic studies reviewed in [JeLe94], [InLe04], 
and neurobiological studies of neural mechanism involved in
coping with tasks related to orientation in physical space (see [CuIt08], [CuItVCYo13]
an brief survey for mathematicians [Ma15].)

\smallskip

In the remaining part of the introduction we will give a very short list of several
approaches to description of  ``meaning'' using geometric/topological representations and/or metaphors.

\medskip

{\bf 1.1. Semic axes.}  In the following it is essential to keep in mind that  core ``meanings''
are generally assigned not to ``words''  but to ``lexemes''. According to [Me16], p.~240,
lexeme is    ``a word taken in one well defined sense -- more precisely,
a set of all word forms and analytical form phrases that differ only by inflectional significations.''
\smallskip
Example ( [Me16], p.~135): lexeme TAKE${}_{(V)}$ includes the following lexical items:
{\it take, takes, took,  taking, \dots , have taken, has taken, \dots, have been taken,\dots }
\smallskip
The tag  {\it (V)} here means that our lexeme refers to the word "take" understood as
 {\it a verb}  rather than {\it a noun.}

\smallskip

When one extracts a vocabulary of lexemes from a dictionary of words, one
should do ``stemming''  (extracting roots of words), ``tagging'' etc., cf. a more detailed description
 in Sec.~3 of [TuPa10].

\smallskip

We will allow ourselves the use of  the term ``word''  in place of ``lexeme'' 
when it cannot lead to a confusion.
\smallskip
The approach to encoding of meaning, or ``sense'' of lexemes,
briefly surveyed in [Gui08], starts with postulating a  list of ``semes'' such as {\it animate,
inanimate, actor, process} etc.
\smallskip
 The meaning is specified by listing a subset of semes.
 \smallskip
 In the respective geometric picture, $N$ semes are represented by basis vectors $e_i$, $i=1,\dots ,N$,
 of $\bold{R}^N$, and meanings are represented by (a subset of) vertices of the unit
 cube $[0,1]^N$.  P.~Guiraud actually prefers the ``bisemic'' description, in which
 meanings are represented by a subset of vertices of $[-1,1]^N$. Sign changes of
 basic coordinates represent the complementarity relations such as in animate/inanimate.
 \smallskip
 A qualitative weakening of the bisemic model allows meanings to be represented
 by points in $\bold{R}^N$ that are localised near the boundary of the unit cube,
 but not necessarily  coincide with its vertices. A nice illustration is given on p.~59 of  [G\"a14].
 It represents bisemes in a two--dimensional ``emotional space'' $\bold{R}^2$
 whose bisemic axes  represent dichotomies {\it pleasure/displeasure} and
 {\it high/low} whereas, say, the quadrant ``low pleasure" accommodates
 lexemes {\it content, serene, calm, relaxed, sleepy}.
 
 \smallskip
 
Some of the largest subsets of the space of meanings that can accommodate, say,   path of a narrative,  might encode notions related to

-- senses: vision, hearing, feelings, time, space \dots

-- some subregions like ``far away -- near'' , ``quiet -- loud'',
``past -- future''

-- regions related to ``me'', to ``other people'' , ``unrelated to humans'', etc.

\smallskip

What is important is that we should construct this semantic space
at first in a way maximally independent of the ``natural language''
we choose, and that it will widen at each stage of construction in order
to accommodate new words, sentences, languages etc. 
 
 \medskip
 
 {\bf 1.2. Semic axes and neural encoding of place field recognition.}  
  We  want to derive from  semantics of a natural language
a structure encoding it that would be a space covered by
subsets, say, $U_i$.  (Some) non-empty finite intersections should correspond
to words or short sentences, paths through this space should correspond to
texts. 
\smallskip
A nice example of this is provided in [Li], together with a picture representing
symbolically two different subsets of semantic space in two possible mutual
relationships: (i)  inclusion of one in another,  and (ii) non--empty intersection
without inclusion. 

\smallskip

This picture illustrates  the difference between usages of words {\it which}
and {\it that} in the following two sentences:
\smallskip
Correct use of {\it that}: { \it ``Tiffany likes shoes that are expensive''.}
\smallskip
``The set of things called shoes includes both expensive and inexpensive shoes, so when we say `that are expensive,' we are talking only about a subset of the set of all things called shoes.''
\smallskip

Correct use of {\it which}:  {\it ``Tiffany likes emeralds, which are expensive''.}
\smallskip
``The set of things called emeralds are all expensive, so the clause `which are expensive' talks about the whole set of emeralds. There is no inexpensive subset of emeralds. `Which are expensive' simply gives you additional information about this whole set'' ([Li]).

\smallskip

This basic picture representing meanings by domains in the space of meanings
and the relationship of intersection/inclusion between  the respective domains
fits very well the studies aimed to  the understanding how brain copes with multiple tasks of
orienting and navigating in the world, cf. [CuIt08], [CuItVCYo13], [Yo14], 
and references therein.
\smallskip
The brain of an animal must be able to reconstruct, say,
a map of its environment and its current position in it, using only the action
potentials (spikes) of the relevant cell groups.
In laboratory experiments 
it is found that stimuli related to the positions are naturally divided into groups, and with each group a
certain type of neural activity is associated.
In [CuIt08] and [Yo14], it is postulated that a given domain of stimuli can be
modelled via a topological, or metric stimuli space $X.$ Furthermore, brain reaction
to a point in $X$ is modelled by spiking activity of certain finite set of neurons $NX$. The list of subsets of $NX$ consisting
of subsets whose neurons can be activated simultaneously, corresponds to a certain
covering of $X$. Thus this covering can be described by a binary code, and relations of intersection/inclusion
between domains coincide with the relations of intersection/inclusion between
the respective code words. For more details, see [Ma15].
 
 \medskip
 
{\bf 1.3. Meaning--Text model.} In the model of semic axes, there is one intrinsic
source of incompleteness: as P.~Guiraud says ([Gui68], p.~157), the lexical units
(corresponding to vertices of $[-1,1]^N$)  {\it ``must in addition be associated in syntagms, each one of them constitutes a `sense'. But there again we must setup rules for combinations, for the sense supposes that certain syntagms are permitted, other excluded.''}  The difference between {\it which} and {\it that} discussed above is precisely
an example of such syntagms.

\smallskip

This problem is very systematically addressed in the model ``Meaning--Text'' which  
I.~Mel'$\roman{\check{c}}$uk
and his collaborators have been developing for several decades: see [Me16] for its most recent summary
and further references.
\smallskip
In this model, meaning of a text of language $\bold{L}$ {\it ``is exclusively [\dots] linguistic meaning''}
that can be extracted {\it ``only on the basis of the mastery of $\bold{L}$, without
the participation of common sense, encyclopaedic knowledge, logic etc.} ([Me16], Sec. 3.2.2).
\smallskip

On the other hand, geometry/topology figures in this model mainly as a tool for producing graphs
of various levels of linguistic representation. Each such graph consists of several vertices,
certain pairs of which are connected by edges. Moreover, both vertices and edges are
additionally marked. For example, on the level of the surface--syntactic structure,
a sentence is represented by the graph, whose vertices are marked by lexemes
corresponding to the words in this sentence, and by additional information encoding
the passage from the lexeme to the word. Edges of this graph are marked by technical
terms expressing syntactic relations between the respective pair of words.

\smallskip

Below we illustrate this principle by presenting the surface--syntactic graph of the
first line of  a sonnet by Michelangelo. We are very grateful to  I.~Mel'$\roman{\check{c}}$uk
who produced for us this graph and allowed us to reproduce it here.

\medskip
\epsfxsize=7cm
\epsfbox{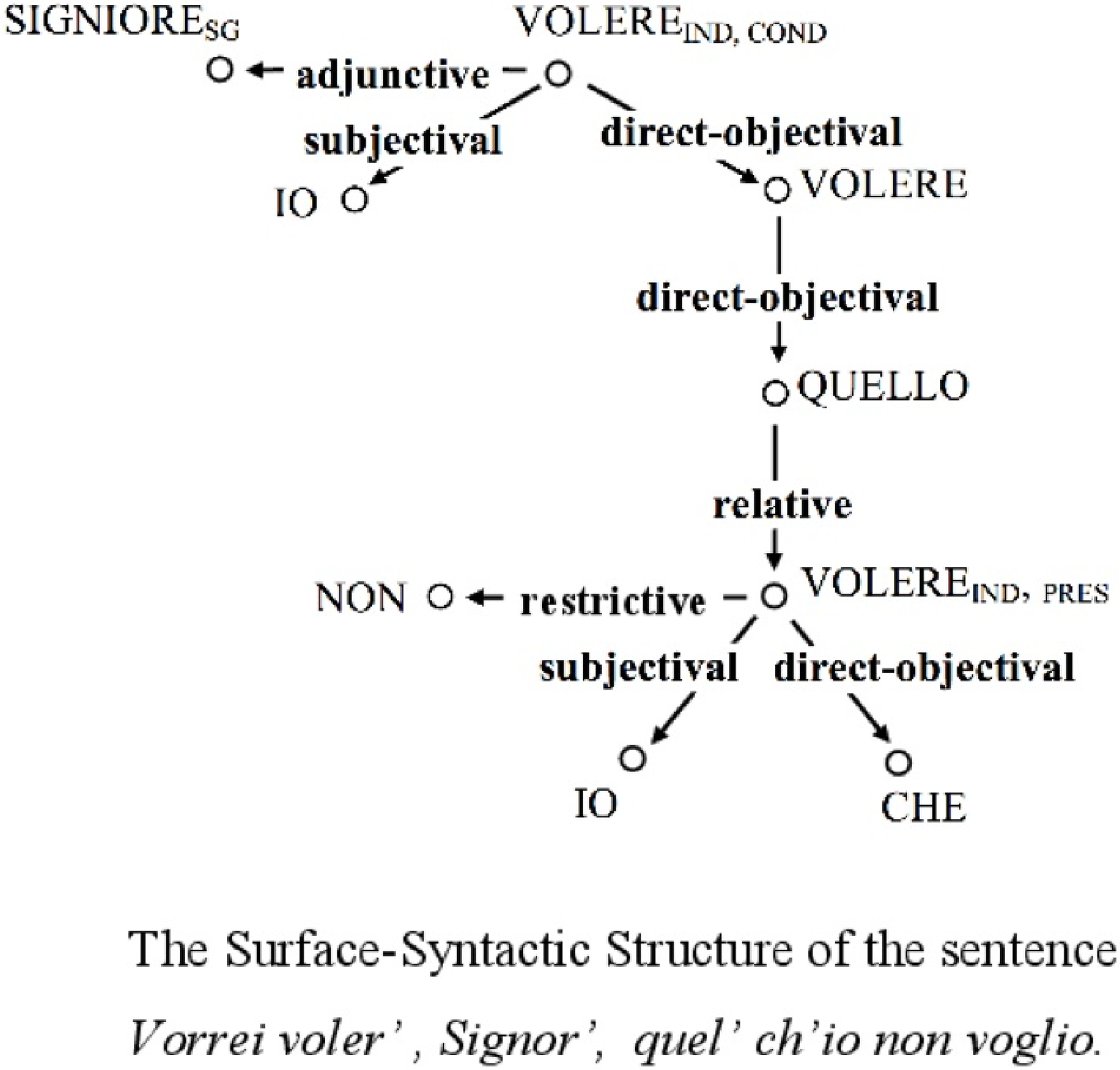}

\bigskip

{\bf 1.4. Neurolinguistic data.} There exists a large body of neuroimaging studies
of production and perception of spoken language. In the survey  [InLe04], the reader will find descriptions 
of methodology used and results obtained in  {\it ``the enterprise of relating the function components
of word production, such as lexical selection, phonological code retrieval, and syllabification,
to regions in a cerebral network''} ([InLe04], p.~102.) 

\smallskip

An illustration of segment of lexical production network ([JeLe94], p. 826) shows fascinating parallels
with the Meaning--Text model.

\smallskip

Due to the vastness of  semantic space needed to accommodate all meanings
expressible in a natural language, direct comparison with the neural encoding of 
place field recognition as in [CuIt08] is not yet feasible. However, the development 
of new methods of studying and collecting databases of results allows us to hope 
that such comparison will become possible. In this paper, we try to contribute 
some mathematical tools that may be useful for this endeavor.

\medskip

{\bf 1.5. This report.}  Most of the approaches discussed above directly
appeal to the linguistic intuition and communicative experience of scientists, experimenters,
and participants in experiments. Information obtained by the respective methods
should be considered as local data about semantic space, and/or about short paths in it.

\smallskip

On the other hand, if we want to obtain mathematical models of topology of  semantic spaces
and of longer routes in such a space, expressed by texts of the size, say, of a chapter in
``War and Peace'', we may turn to the statistical natural language processing.

\smallskip

Then, in the first approximation, a text becomes a point in the space of paths in the semantic space, and we discuss
here approaches to  studying the topology of such spaces appealing
mostly to the data about frequencies of  lexemes and other text fragments
taking in account their linear ordering in the text. Semantics of such fragments
as it is represented by dictionaries and experiments is thus put aside to a certain degree,
although not fully.

\smallskip

In the main body of this article, we will  describe some mathematical 
tools that can be used for the introduction of  ``time dimension'' in the study
of texts.  They will refer to the geometry of real projective spaces and
real Grassmannians. Passage from texts to  the relevant geometry
is based here on the  Vector Space Models of semantics (VSM) surveyed
in [TuPa10], and we will briefly explain  this model for further use.

\smallskip

In Section 2 we discuss how the frequency matrix of the
VSM approach, that counts occurrences of lexemes in contexts in a
given corpus of texts, determines a point in a Grassmannian.  We show
that, in the case of a large vocabulary of lexemes and a smaller number of contexts, 
the condition that the resulting point lies in the positive Grassmannian provides a 
geometric test for the property that a choice of lexemes gives a good semantic 
disambiguation of the contexts. A similar condition holes in the case of a 
small number of lexemes, where one wants to test if a set contexts would 
disambiguate the words semantically. This geometric viewpoint takes into
account the fact that contexts come with a specific ordering by occurrence in a text. 

\smallskip

In Section 3 we discuss other geometric models associated to the
frequency matrices of the VSM approach, which also takes into account
the specific ordering of contexts in a text. We assign to a text a piecewise 
geodesic path of points in a projective space. Instead of measuring semantic
relatedness in terms of angle distances between the semantic vectors of
the frequency matrices, as it is customary to do in Natural Language Processing,
we compare the paths in an ambient projective space through a geometric
distance function between (geodesic) polygonal curves, which is known
to be computable in polynomial time. In a variant of this construction, we
also consider assigning to the frequency matrix a point in a flag variety,
where the flag corresponds to the span of successive semantic vectors
for the successive contexts ordered by occurrence in a text. Again semantic
similarity can be measured in terms of the geodesic distance in the flag
variety, with respect to its natural metric as a quotient of Lie groups. 

\smallskip

In Section 4 we consider the case where lexemes are grouped together
according to some semantic axes, either by explicit semantic tagging
(supervised learning) or just by grouping together lexemes with
similar occurrences in contexts (unsupervised learning). In both cases,
we describe the process of passing from frequency matrices for a given
corpus of text, computed with respect to a dictionary of lexemes, to
density matrices with respect to a semantic dictionary, where
identification of lexemes by semantic criteria has already occurred.
When we view the frequencies as determining points in Grassmannians,
we can view geometrically this operation as a projection between
two Grassmannian. The question of whether one can avoid loss of semantic
information in this process, when applied to a given collection of texts,
is then interpreted in terms of whether the points corresponding to
these texts lie on a subvariety of the Grassmannian that can be 
isomorphically projected to the other Grassmannians. A similar 
condition arises when we assign to a given text a piecewise geodesic
path in a projective space as discussed in Section 3. 

\smallskip

In Section 5 we connect the geometric setting described in
the previous section with the point of view of persistent topology. 
According to our previous construction, a large corpus of texts
determines a corresponding set of points in an ambient
Grassmannian, where we assume that the same fixed dictionary of
lexemes (or semes) is used to analyze all texts in the corpus. 
We then show that one can identify more refined forms of 
semantic relatedness between these points. These are topological in 
nature and arise from constructing Vietoris--Rips simplicial complexes
at varying scales, associated to the set of points in the ambient 
variety and computing their persistent homology. We discuss
possible relations to the use of persistent topology in the
theory of neural codes.

\smallskip

In Section 6 we show that the Latent Semantics technique 
for dealing with very sparse frequency matrices in the VSM
approach, which identifies lower dimensional subspaces 
(latent meanings) through singular value decomposition, can
be interpreted in terms of the geometry of Grassmannians
described in Section 2, as a Riccati flow on the ambient
Grassmannian. 

\smallskip

In Section 7 we discuss how to implement, in our geometric
setting, a model analogous to G\"ardenfors' ``meeting of
minds", where common meaning between different
users communicating with one another is achieved as
via a fixed point problem in a convex semantic space.
We suggest that a similar idea can be implemented in
our setting if different users come to somewhat different
semantic interpretations of a given texts, on the bases
of semantic interpretations based on other texts available 
to them, under the assumption that users have access
to different (partially overlapping) corpora of texts.
We then describe the procedure of ``meeting of minds"
as the construction of a geodesic barycenter in the
ambient geometric space of the distribution of points
obtained by the users, possibly weighted according to some
measure of ``reliability" of the different corpora used for
semantic interpretation. 

\smallskip

In Section 8 we discuss how to compensate for the
fact that the frequency distribution for words in a
dictionary is skewed towards the more frequent
and less semantically significant words according
to Zipf's law.

\bigskip

\centerline{\bf 2. Vector Space Models of semantics}

\medskip

{\bf 2.1. Texts and their processing.} A concrete VSM  starts with a large corpus of natural language
texts and produces from it a matrix of numbers (frequences).  The intermediate steps of this production are subdivided into
two groups: {\it  (i) linguistic processing}, and {\it (ii) statistic processing}.

\smallskip

For us, linguistic processing results in the creation of the relevant {\it vocabulary of lexemes}
where we understand ``lexeme'' as in 1.3 above.  Each text is also represented as  a sequence
of the relevant lexemes, although from the description of [TuPa10] it becomes clear
that at least some fragments of it are modelled by their surface--syntactic structures
in the sense of  Meaning--Text model.
\smallskip
We accept this as a reasonable approximation
to the procedures described in Sec.~3 of [TuPa10].

\smallskip

Statistic processing, as we mentioned, produces a (normalised) {\it matrix of frequencies,}
see [TuPa10], Sec.~4.

\smallskip

In the typical case called ``the term--document matrix'' in [TuPa10], 
rows of the matrix are labelled by lexemes (``terms''), whereas 
columns are labelled by texts in our collection.

\smallskip

In another typical case called ``the word--context matrix'' ([TuPa10], Sec.~2.5),
the texts, already  at the stage of linguistic processing, are represented as a union 
of ``contexts''.  Here again the rows of matrix are labelled by lexemes, whereas
columns are labelled by contexts.

\smallskip

Finally, matrix entries as we mentioned characterise correlations between
the  lexemes and text/contexts. We will treat in more detail some cases below, and address the question of 
``smoothing''.

\medskip

{\bf 2.2. ``Time dimension'' and other linear orderings.} Any vocabulary
of lexemes, or contexts, must be in the final count also presented
as linearly ordered dictionary. This ordering might be totally irrelevant
to the situation under study (as e.g.~alphabetic ordering). It can take
into account the order of first appearance of the respective lexeme
in the text. Finally, it can be a Zipf's--like ordering according
to diminishing frequency rate.

\smallskip

A considerable part of statistical characteristics of a VSM does not depend
of the chosen orderings (although the mode of their usage
might depend on it).  However, for the purpose of our paper
this might become essential, and we will pay due
attention to it.

\medskip

{\bf 2.3. Notation and assumptions.} We will consider word--contexts matrices described above
in one of  two possible extreme subcases.
 
 \smallskip

(A). {\it Large vocabulary case.}  In this setting, we assume  that our
vocabulary of lexemes  is sufficiently large and includes at least all the lexemes that appear in the texts
(excluding words with large occurrences in all contexts 
such as ``and" or ``the" in an English text that are semantically less informative).
Moreover, we assume that the size of the vocabulary is large compared to the number of
contexts in the texts. In this case, one aims at selecting from the large
dictionary choices of words that best represent the given contexts semantically.

\smallskip

(B). {\it  Information retrieval case}. In this case we consider a vocabulary that
is small compared to the number of contexts, as would be the case with a 
choice of words used in a query. In this case one aims at selecting among the
various contexts in a given corpus those that best match semantically the chosen
words in the query.
 
\medskip

 Let $D$ be our vocabulary, and $M=\# D$ be the number of lexemes in it.

\smallskip 

A  given text $T$  is then endowed with a set of subtexts called {\it contexts}:  $C(T)=\{ c_1, \ldots, c_N \}$.
Typical examples of contexts are: sentences, paragraphs, or else windows of certain length
around each word/lexeme.

\medskip
{\bf 2.4. Matrix of frequencies.} Following [TuPa10], one produces from these data an
$N\times M$ matrix of frequencies $P=P(T)$ with  entries $p_{ij}$. Here
$p_{ij}$ is the estimated probability (frequency) of occurrence of the word $w_i\in D$ in the
context $c_j\in C(T)$. In the VSM model, one usually considers also
the matrix $X=X(T)$ with entries $X=(x_{ij})$,
$$
 x_{ij} = \max\{ 0, \log\left( \frac{p_{ij}}{p_{i\star} p_{\star j}} \right) \}, 
$$
where $p_{i\star}=\sum_j p_{ij}$ is the estimated probability of the word $w_i\in D$ 
and $p_{\star j}=\sum_i p_{ij}$ is the estimated probability of the context $c_j\in C(T)$.
The condition that $p_{ij}=p_{i\star} p_{\star j}$ corresponds to statistical independence
of word $w_i$ and context $c_j$, while $p_{ij} > p_{i\star} p_{\star j}$ signals the 
presence of a semantic relation between them. 

\smallskip

More precisely, the formula $p_{i*}=\sum_j p_{ij}$ 
gives the frequency of appearance in the text in the case
where contexts do not overlap whereas their union is the whole text. 

\smallskip
In the more general case where contexts may
overlap one still uses the same matrix but
now its entries are the frequencies of appearance across
all contexts (or, equivalently, the frequencies of
appearance in the text, weighted by some
multiplicities that keep track of when a word
appears in the intersection of more than one
context).
\smallskip
If a word is in the intersection of two adjacent
contexts $j$ and $j+1$, then it affects the counting
in both $p_{ij}$ and $p_{i, j+1}$, so $\sum_j p_{ij}$
is still the normalization factor.
\smallskip
 The typical
example of this ``overlapping contexts" method
is the original Shannon 3-gram model: here one
has probabilities (based on frequencies) for
occurrences of 3 words in a row. For example,
one can have a word sequence   a-b-c-d where
the three words  a-b-c have a very high probability
of occurring together, while the probability of
the triple b-c-d  is very low. This suggests that
it is  a-b-c rather than b-c-d that clarifies better
the semantic meaning of the words b and c, and
that the semantic meaning of d will more likely
be clarified by the trigrams that follow like c-d-e
and d-e-f, with the following words e,f,
rather than by b-c-d.

\medskip

{\bf 2.5. Large Vocabulary case}. Here we have  $N\leq M$. The dictionary 
$D$ includes (at least) all the (relevant) lexemes that occur in the text $T$, 
and the number of contexts in which the text $T$ is subdivided is  smaller 
than the number of words in the dictionary. 

\smallskip

The {\it Statistical Semantics Hypothesis} states that statistical patterns of
word usage in texts determine their semantical meaning, and in particular
that (parts of) text that have similar vectors in the above frequency matrices
also have similar meanings. 

\smallskip

Let $r=rank(P)$ be the rank of the matrix $P(T)$. In the case of large dictionary,
we have $r\leq N$. Under the Statistical
Semantics Hypothesis, the rank $r$ measures the largest
number of words and contexts that the text $T$ disambiguates semantically.
Namely, the linear dependence of frequency vectors is interpreted as revealing
the presence of underlying semantic relations. When $r=N$, all the contexts
in $T$ have a choice of corresponding words that they semantically disambiguate.

\smallskip

In the case where $r=N$, the matrix $P(T)$ of a text $T$ determines a point $p(T)$ in the
real Grassmannian $Gr(N,M)$ of $N$-planes in real Euclidean space $\R^M$. Similarly,
if $rank(X(T))=N$, the matrix $X(T)$ determines a point $x(T)\in Gr(N,M)$. For
simplicity, we argue about the matrix $P(T)$. When not otherwise stated, the
same will apply to $X(T)$.
Let $\cM=\cM(T)$ be the set of subsets of $\{ 1,\ldots, M \}$ of cardinality $N$,
such that the determinant of the corresponding minor is $\Delta_I(P(T))\neq 0$.
The set $\cM$ determines a matroid stratum $\cS_\cM \subset Gr(N,M)$, with
$P(T)\in \cM$.

\smallskip

In the case $N\leq M$, instead of working with a fixed (large) dictionary $D$
for all texts, it is convenient, given a text $T$, to discard all the words in $D$ 
that do not appear anywhere in $T$, as the text does not have any relevance
for those words. Thus, we can assume that $D=D(T)$, with $\# D(T)=M(T)$
is the list of words that appear in $T$ (with a suitable stop list). 
A text $T$ has a linear ordering, which induces an ordering on the set
$D(T)$ that lists words in order of apparition in $T$. We identify $D(T)$ with 
the set $\{ 1, \ldots, M(T) \}$ using this ordering. Similarly, the set $C(T)$
of contexts is also ordered by how they are ordered in the text $T$, and
we identify $C(T)$ with $\{ 1, \ldots, N(T) \}$ using this ordering. The order
of apparition of words in the text $T$ is relevant to the semantic interpretation
of the text, as the first occurrence of a word is the first instance where a semantic
interpretation for that word is required.  

\smallskip

Consider then the set of subsets $I=\{ i_1, \ldots, i_N \}$ of $[M]:=\{ 1, \ldots, M \}$
with $i_1< i_2 < \ldots < i_N$. These correspond to choices of words $w_{i_1}, \ldots,
w_{i_N}$ in $D(T)$, such that the order of apparition of these words in the text $T$ is respected, 
and we consider the frequency vectors $P_{i_k}:=(p_{i_k,j})_j$ for the occurrence of the word $w_{i_k}$
in the context $c_j$. We consider the Gale ordering on these subsets $I$. 
Namely, two such subsets $I=\{ i_1, \ldots, i_N \}$ 
and $J=\{ j_1, \ldots, j_N \}$, with $i_1< i_2 < \cdots < i_N$ and $j_1< j_2 < \cdots < j_N$,
we have $I \leq_G J$ iff $i_1\leq j_1$, $i_2\leq j_2$, $\ldots$, $i_N\leq j_N$. The Gale
ordering corresponds therefore to the relative position of words $w_{i_k}$ and $w_{j_k}$ 
in the dictionary $D(T)$ according to first apparition in $T$.

\smallskip

The original dictionary $D$ also has an ordering, and therefore the smaller dictionary $D(T)$
also has an induced ordering, which is different than the order of apparition in the text $T$.
One then has some permutation $\sigma \in S_M$, such that the Gale ordering described 
above corresponds to the ordering $I \leq_\sigma J$, namely
$\sigma^{-1} I \leq_G \sigma^{-1} J$. 

\smallskip

The condition that, for one of these subsets $I$, the corresponding minor of the matrix $P(T)$
has vanishing determinant $\Delta_I(P(T))=0$ means that there is a linear dependence
between the vectors $P_{i_k}$, hence under the Statistical Semantics Hypothesis
a semantic relation between the $w_{i_k}$. Thus, the matroid stratum 
$\cS_\cM \subset Gr(N,M)$ containing the point $p(T)\in \cS_\cM$
determined by the text $T$ describes, for the given contexts $c_i$ of the text $T$, 
all the choices of words  $w_{i_k}$, $k=1,\ldots,N$ in the dictionary for which the 
semantic vectors $P_{i_k}$ are independent. This can be seen as the maximal
amount of semantic information that can be extracted from the text and its contexts.

\smallskip

Recall that the positive (or totally non-negative) Grassmannian $Gr_{\geq 0}(N,M)$ is the subset
$Gr_{\geq 0}(N,M) \subset Gr(N,M)$ of matrices $A$ such that for all $\Delta_I(A)\geq 0$, for
$I$ as above. The intersections of the matroid strata with the positive Grassmannians
$\cS_\cM^{\geq 0}=\cS_\cM\cap Gr_{\geq 0}(N,M)$ are cells, the {\it positroid cells} of [Pos06].

\smallskip

In particular, the condition that the point $p(T)$ lies in the positroid cell $\cS_\cM^{\geq 0}$,
that is, that all $\Delta_I(A) > 0$, for all $I \in \cM$, is equivalent to the existence of continuous
paths $\gamma_I$, for each $I \in \cM$, where $\gamma_I(0)=P(T)$ and $\gamma_I(1)$ is
a matrix where the $I$-minor is the identity, and for all $t\in [0,1]$ one has 
$\gamma_I(t)\in \cS_\cM^{\geq 0}$. This condition can be regarded as expressing
the fact that the choice of words $w_{i_1}, \ldots, w_{i_N}$ for the contexts $c_1, \ldots, c_N$
of the text $T$ contains a maximal amount of semantic information. Indeed, the case where
the corresponding minor would be the identity, would correspond to a case where the word
$w_{i_k}$ is entirely specified semantically by the context $c_k$ and by none of the other
$c_j$ with $j\neq k$.

\medskip

{\bf 2.6. Information Retrieval case.} We now focus on the other case mentioned
above, the ``information retrieval" setting, where we have $N\geq M$,
that is, where the list of words is, for example, 
the list of words in a query, and one wants to locate texts, or contexts
within a text, that are semantically most relevant for that query. 
In this case, we can assume that the number of words searched is 
no greater than the number of contexts.

\smallskip

The setting is similar to what we described before, except that we
now consider the case where the matrix $P(T)$ determines a point
in the Grassmannian $Gr(M,N)$. The minors $I=\{ i_1,\ldots, i_M\}$
correspond to choices of contexts $c_{i_k}$ in the text $T$ in response 
to a query given by the words $w_k$. As before the condition
$\Delta_I(P(T)) > 0$ corresponds to those assignments of a context
to each word of the query that best matches it semantically.

\medskip

{\bf 2.7.  Literary texts and their statistical processing.} D.~Yu.~Manin in his
article [Man12] suggests that literary texts (prose/poetry) require
qualitatively different methods of statistical processing in order
to make explicit what puts them apart from texts produced in ordinary speech.

\smallskip

Here we only mention a different kind of contexts
 used there ([Man12], p.~286).
 
\smallskip

Namely, a context in his sense is a fragment of text with a blank, a hole where different words might occur, like  ``a--*--b". This would allow one to extract statistical data allowing one to say that
  ``words x and y often occur in the same contexts". Presumably, this fact would then reflect
 semantic relationships between x and y. 
  
 \smallskip
  
  In the limiting case where x can occur in all the same contexts as y, 
  and with the same frequencies, that would mean that x and y are exact synonyms. 
  Or, if x can share contexts with u and v, but u and v do not share contexts, then they probably 
 represent two very distinct meanings of x. 

\smallskip

In this paper, we do not try to study semantic spaces and paths in them relevant to
this approach. We only mention that it might be a very interesting project.

\vskip1cm

\centerline{\bf 3. Projective Spaces and Flag Varieties}

\medskip

We describe here two variants of the construction above,
aimed at encoding more explicitly the fact that a linguistic
text has an ordered linear structure that is crucial to its
semantic interpretation. We propose two modifications of
the geometry described above that better encode this fact.
One is based on regarding a text subdivided into contexts,
as a collection of points determining a path in a projective
space, rather than as a single point in a Grassmannian.
The second is in terms of points in a flag variety. 

\medskip

{ \bf 3.1. Texts as Paths in Projective Spaces.} Here we again consider
the case where we have some fixed large vocabulary $D$ of
lexemes of size $M=\# D$, which contains at least all the words
in the given text $T$. We also subdivide the text into contexts $c_k$,
as before, but we do not necessarily assume that the total number $N$
of contexts is smaller than $M$. Indeed, in this setting
we could be dealing with a large corpus of texts and a large 
number of contexts. We again consider the semantic vectors
$P_k(T)=(p_{ik})_{i\in D}$ that collect the probabilities (frequencies)
of occurrence of words $w_i \in D$ in the context $c_k$ of $T$.
We regard each $P_k$ as determining a point $p_k$ in the
projective space $\P^{M-1}\simeq Gr(1,M)$. Thus, a text $T$ here
corresponds to an ordered $N$-tuple of points in $\P^{M-1}$, where
$N$ is the number of contexts. We can think of this collection
of points as an oriented path by drawing geodesic arcs between 
consecutive points. We denote by $\Gamma(T)$ the resulting path 
associated to a text $T$.

\smallskip

Given different texts $T$ and $T'$, the comparison at the
level of semantic vectors can be performed, in this setting,
by computing the distance between the corresponding paths 
in the same ambient $\P^{M-1}$. This can be computed as the
Fr\'echet distance between the two polygonal curves. The
latter is defined as the infimum over reparameterizations 
by $[0,1]$ of the maximum over $t\in [0,1]$ of the distance
between corresponding points
$$ \delta(\Gamma(T),\Gamma(T'))=\inf_{\gamma,\gamma'} \max_{t\in [0,1]} d_{FS}(\gamma(t),\gamma'(t)), $$
where $\gamma: [0,1]\to \Gamma(T)$ and $\gamma': [0,1]\to \Gamma(T')$
are parameterizations of the two curves by $[0,1]$, and $d_{FS}(x,y)$ is the Fubini-Study
metric on $\P^{M-1}$. The Fr\'echet distance for polygonal curves
is computable in polynomial time ([AlGo95]). 

\medskip

{\bf 3.2. Texts as points in flag varieties.} Another way to keep track of the linear
ordering of contexts in a given text is by building larger subspaces, as more
and more contexts in the given texts are encountered in a linear reading 
of the text. Thus, if $P_k(T)=(p_{ik})_{i\in D}$ are the semantic vectors as
above, one considers the vector spaces $V_k=span\{ P_j \,:\, j=1,\ldots, k \}$.
The spaces $V_1\subset V_2 \subset \cdots \subset V_N$ form a 
flag in $\R^M$. We denote by $F(d_1,\ldots, d_\ell)$ the flag
varieties of flags $W_1\subset\cdots W_\ell$ with $\dim(W_k/W_{k-1})=d_k$.
We associate to a text $T$ the point of the corresponding
flag variety $F(1,\ldots, 1, M-N)$ determines by the flag 
$V_1\subset V_2 \subset \cdots \subset V_N$ with
$V_k=span\{ P_j \,:\, j=1,\ldots, k \}$.

\smallskip

The natural Fubini--Study metric on projective spaces has an analog
for Grassmannians and flag varieties. It is obtained from the curvature form
of the first Chern class of the determinant line bundle of a hermitian
vector bundles ([Dem88]), or else by considering these
varieties as quotients of $SU(n)$ by subgroups, with the metric induced
from the bivariant metric of $SU(n)$ ([Gri74]). Thus, one can compare texts
viewed as points in Grassmannians or in flag varieties, by measuring 
their distance with respect to this metric. 

\bigskip

\centerline{\bf 4. From Lexemes to Semantic Dictionaries}

\medskip

We now consider a setting where, instead of a ``lexemes dictionary" $D$ of words,
one passes to a ``semantic dictionary" $S$ where lexemes are grouped together
according to some semantic description. This can happen in two different ways,
based on supervised or unsupervised learning.
\smallskip
(a)  {\it Supervised Learning.} In this case, also referred to as ``sense tagging"
(see [MaSch99]), lexemes are grouped together into semantic categories by
assigning appropriate tagging. In this setting, the type of question we look at
is to what extent the information contained in the semantic vectors computed
for the initial lexical vocabulary still retains the correct information when
passing to a quotient that corresponds to the identification by semantic
categories. 
\smallskip

(b)  {\it Unsupervised Learning.} In this case, sense tags are not assigned,
so that one cannot identify directly the corresponding semantic categories,
but one can still obtain a ``sense discrimination" by grouping together 
words into unlabelled groups using the information contained in the
semantic vectors. In this setting, we will show that the resulting
grouping can be studied in terms of {\it persistent topology}  ([Ca09]).

\medskip

{\bf 4.1. Supervised Learning.} We  consider the case where we
associate to texts $T$ points $p(T)$ in a Grassmannian (either
$Gr(N,M)$ or $Gr(M,N)$ depending on relative size of vocabulary
and contexts). We consider the case $N< M$. The other possibility can be
treated similarly. 

\smallskip

We want to consider also the case where we 
deal not with a single text $T$ but with a corpus consisting
of several texts. In this case, we need to assume that the
vocabulary $D$, with $M=\# D$, is large enough to include all
words that occur in all the texts of the corpus. Moreover, if we
choose an ordering of the dictionary, as discussed previously,
by order of apparition in a text, we can extend the order
to the whole corpus, by choosing an order in which the
different texts in the corpus are looked at. For the model with
points in Grassmannians, or in flag varieties, we consider
the case where the number $N$ of contexts is fixed across
all texts in the corpus. In the more general case where
the number $N=N(T)$ varies across texts, we will be working
with the model in which texts determine a sequence of points
and an oriented polygonal path in a fixed projective space.
In both cases, the question will be the behavior of the
locus (in the Grassmannian, flag variety, or projective space)
determined by the semantic vectors of all texts in the corpus,
under a projection map that corresponds to passing 
from the lexical to the semantic dictionary. 

\medskip

{\bf 4.2. Points in Grassmannians and Flag Varieties.} 
At the level of the matrix $P(T)$ and the corresponding point $p(T)$ in the
Grassmannian $Gr(N,M)$, 
one can view the operation of passing from the lexemes in $D$ to the
semantic categories in $S$ as the effect of a projection 
$\pi_{M,M'}: Gr(N,M) \twoheadrightarrow Gr(N,M')$, where $M'\leq M$ is the
size of the set of semantic categories considered, $M'=\# S$. 

\smallskip

We regard a corpus $\cC =\{ T \}$ of texts $T$ as a discrete sampling of a 
subvariety of the Grassmannian $Gr(N,M)$, under the hypothesis that
the number of contexts is fixed and the size of the dictionary $D$ is also
fixed for all $T \in \cC$. We denote by $\Pi_\cC=\{ p(T) \}_{T\in \cC}$ the
finite set of points on $Gr(N,M)$ corresponding to the texts in the corpus.
Given the finite set $\Pi_\cC$, we consider possible algebraic 
subvarieties $X_\cC\subset Gr(N,M)$ that  interpolate the points $p(T) \in \Pi_\cC$,
namely algebraic subvarieties $X_\cC$ of $Gr(N,M)$ with $\Pi_\cC \subset X_\cC$. 

\smallskip

We recall some results about projectability of subvarieties of Grassmannians,
see [ArRa05]. A subvariety $X\subset Gr(N,M)$ is $k$--projectable, for some
$0\leq k\leq N-1$, under $\pi_{M,M'}: Gr(N,M) \twoheadrightarrow Gr(N,M')$ if any two
$N$--planes in the image of $X$ only meet along linear spaces of dimension
less than $k$. The case $k=N$ corresponds to $X$ being isomorphically
projectable to $Gr(N,M')$. Note that $k$--projectability also implies that no two 
$N$--planes in $X$ can intersect in dimension greater than or equal to $k$.

\smallskip

If the variety $X_\cC$ associated to a corpus $\cC$ of texts is $k$--projectable
to $Gr(N,M')$, this means that the $N$-planes given by the images 
$\pi_{M,M'}(p(T))$ and $\pi_{M,M'}(p(T'))$ of any two points $p(T)$, $p(T')$,
with $T,T'\in \cC$, will intersect in at most a  $(k-1)$-dimensional space. 

\smallskip

The size of the intersection between the $N$-planes of $T$ and $T'$ is
a measure of dependence between the respective semantic vectors,
hence of the semantic relatedness of the two texts. If in the variety $X_\cC$
every two $N$--planes intersect in dimension less than $k$, but the variety
is not $k$-projectable under $\pi_{M,M'}: Gr(N,M) \twoheadrightarrow Gr(N,M')$,
this means that there is loss of semantic information in the matching of
words (and their semantic categories) to contexts in the texts of the corpus. 

\smallskip

There are strong algebro--geometric restrictions on $k$-projectable
varieties. For example, it is shown in [ArRa05] that the Veronese embedding
of $\P^n$ is the only variety in $Gr(d-1,dn+d-1)$ that can be projected to
$Gr(d-1,n+2d-3)$ so that any two $(d-1)$-planes meet in at most one point.

\smallskip

We have only discussed here the case where we associate 
texts to points in Grassmannians. The case of points in flag
varieties is similar, with similar questions about $k$-projectable
subvarieties.

\medskip

{\bf 4.3. Paths in Projective Spaces.} 
We then consider the case where the size $N=N(T)$ of contexts in a text is varying
with $T\in \cC$. In this case, instead of working with texts defining points in a Grassmannian,
it is more convenient to adopt the viewpoint where texts determine polygonal paths in 
a projective space $\P^{M-1}$ with $M=\# D$ the size of the dictionary. In this case, we
are looking at a similar question about $k$--projectable subvarieties in projective spaces.

\smallskip

More precisely, we consider again algebraic subvarieties $X_\cC$ of $\P^{M-1}$
that contain all the paths $\Gamma(T)$ for $T\in \cC$. As a weaker condition, we can just assume that
the variety $X_\cC$ contains the set of points $\Pi_\cC=\{ p_k(T)\,:\, T\in \cC, \, k=1,\ldots, N(T)\}$.
If $X_\cC$ is also geodesically complete, then it contains also the paths $\Gamma(T)$.

\smallskip

We then consider a projection $\pi_{M,M'}: \P^{M-1} \twoheadrightarrow \P^{M'-1}$ that corresponds to
performing some identification of the vocabulary by grouping lexemes according to a
choice of semantic categories, with $M'=\# S$.

\smallskip

We are then looking at the problem of whether it is possible to project isomorphically
a subvariety $X_\cC$ of $\P^{M-1}$ that contains the points $\Pi_\cC$
(and possibly the collection of paths $\Gamma(T)$) to the quotient $\P^{M'-1}$.
Again, there are strong restrictions on the existence of such isomorphically
projectable subvarieties. For example, it is shown in [Ar01] that the only $n$-dimensional
variety that can be isomorphically projected from $Gr(1,2n+1)$ to $Gr(1,n)$ 
is the Veronese variety, that is, the embedding of $\P^n$ in $Gr(1,2n+1)$
via $O_{\P^n}(1)^{\oplus d}$.

\smallskip

When the variety $X_\cC$ is not isomorphically projectable from $\P^{M-1}$ to
$\P^{M'-1}$, there is some loss of information in the semantic vectors, when
the identification of words according to semantic tags is performed. In such cases,
which will be typical in view of the very restrictive condition of isomorphic projectability,
one can describe the effect of the identification on semantic vectors by analyzing
the change of topology in the polygonal path $\Gamma_\cC =\cup_{T\in \cC} \Gamma(T)$.
We describe ways of approaching computationally such topology changes. 

\medskip

{\bf 4.4. Persistent Topology.} It was understood in recent years that clusters of data 
points can exhibit interesting topological structure that can be useful in analyzing
large data set, see [Ca09] for a general introduction and overview of the field of
persistent topology. Applications of persistent topology to Linguistics were
recently discussed in [PorGhGuCLDMar15]. 

\smallskip

Given a set $\Pi$ of points in a metric space,
one considers a family of simplicial complexes, parameterized by a real number
$\epsilon >0$, the so called Vietoris--Rips complexes $R(\Pi,\epsilon)$.  Here
the $n$--th term $R_n(\Pi,\epsilon)$ is the vector space spanned by all the
unordered $(n+1)$-tuples of points in $\Pi$ where all pairs have distance
at most $\epsilon$. There are inclusion maps $R(\Pi,\epsilon_1)\hookrightarrow R(\Pi,\epsilon_2)$
when $\epsilon_1< \epsilon_2$. These induce maps in homology $H_n(R(\Pi,\epsilon_1))\to
H_n(R(\Pi,\epsilon_2))$. In analyzing the dependence on $\epsilon$ of the ranks of these
homology groups one discards as ``noise" those generators that arise and disappear
within a small range of values of $\epsilon$, while one regards those generators that
persist for sufficiently long intervals of values of $\epsilon$, the ``persistent generators",
as signaling the presence of actual structure in the data. 

\medskip

{\it Persistent topology of the set $\Pi_\cC$ in the Grassmannian.} Persistent
topology can also be used to enrich the semantic comparison of different texts,
when we assign to each text in a corpus a point in a Grassmannian or in a
flag variety, as discussed above. The simplest level of comparison would be
to cluster together the points corresponding to the various texts by separating
them into groups according to the relative distances in the ambient metric. The
resulting groups are dependent upon the scale of the neighborhoods of points,
and the number of different groups of semantic similarity
correspond to the rank of the zeroth order persistent homology of the 
Vietoris--Rips complex. Thus, more refined information about how texts
cluster together by semantic similarity is obtained by additionally considering also
the first and higher dimensional persistent homology. 

\smallskip

We consider, as above, a projection $\pi_{M,M'}: Gr(N,M) \twoheadrightarrow Gr(N,M')$
and the image $\pi_{M,M'}(\Pi_\cC)$. In the case where the set of points $\Pi_\cC$
does not fit on an interpolating variety that is isomorphically projectable, we can
analyze the change in the semantic proximity of texts by analyzing the differences
between the persistent topology of $\Pi_\cC$ and of $\pi_{M,M'}(\Pi_\cC)$. This can
be seen by computing the number of persistent generators, in various degrees, of
the homology of the respective Vietoris--Rips complexes. The case of points in
flag varieties can be treated analogously to the case of points in Grassmannians.

\medskip

{\it Persistent topology and paths in $\P^{M-1}$.}  In a similar manner, one
can use persistent topology to analyze syntactic proximity of texts in the 
point of view where we assign to each text in a corpus a path in projective
space $\P^{M-1}$. In this case, we again associate to a corpus $\cC$ a
simplicial complex, where the zero-cells are all the points $p_k(T)\in \P^{M-1}$,
for all texts $T\in \cC$, and all the one-cells are the geodesic arcs connecting
consecutive pairs of points $p_k(T)$ and $p_{k+1}(T)$. 
The higher dimensional skeleta are then constructed as in the Vietoris--Rips complex,
by adding an $n$-dimensional simplex whenever an $n+1$-tuple of points
$\{ p_{k_0}(T_0), \ldots, p_{k_n}(T_n) \}$  where the geodesic distances
between all pairs of these points are less than a fixed scale $\epsilon$.
This may require introducing additional one-cells in the complex.

\smallskip

As in the case of points in Grassmannians and flag varieties, when we
consider a projection $\pi_{M,M'}: \P^{M-1} \twoheadrightarrow \P^{M'-1}$,
we can study the effect of the projection on the persistent topology of
the set of paths $\Gamma_\cC =\cup_{T\in \cC} \Gamma(T)$ and
its image $\pi_{M,M'}(\Gamma_\cC)$, by associating complexes as
indicated above to $\Gamma_\cC$ and $\pi_{M,M'}(\Gamma_\cC)$
and comparing generators of the respective persistent homologies. 

\medskip

{\bf 4.5. Unsupervised Learning.} In the case of unsupervised learning,
a grouping corresponding to ``sense discrimination" is obtained solely on
the basis of the semantic vectors and the position of the corresponding points in
the ambient variety, without any external tagging of words by semantic categories. 
In the setting of unsupervised learning, the grouping together of subsets
of the $M$ lexical dimensions into putative semantic categories is itself
performed solely on the basis of the semantic vectors. A simple way to
search for semantic relatedness in an unsupervised context is to identify
frequent co-occurrences within the same contexts (see Section 2.4 of [TuPa10]). 
Many co-occurrences arise for purely syntactic reasons, but those tend to
be between words that belong to different parts of speech, while co-occurrences 
that carry semantic significance are more often found between words in the same 
part of speech, see [BuHi06], [ChiaBRP90], [SchPe93], [TuPa10].

\medskip

{\bf 4.6. Syntactic dependence of semantic vectors.} Clearly, the
vectors $P_k(T)=(p_{ik})_{i\in D}$, associated to the contexts $c_k$ in a
text, depend on both syntactic and semantic information and there is
a priori no obvious way to distinguish between the dependence on
syntax and on semantics. However, a possible way to make these
semantic vectors more syntax independent would be to consider a
training corpus of different language translations of the same texts, 
with marked matching paragraphs and matching word dictionaries,
and average the semantic vectors $P_k(T,L)$ over the set of languages $L$.
This can be done either by simply averaging the vectors, or else by
considering the corresponding points $p_k(T,L)$, for all languages $L$, in 
the fixed ambient $\P^{M-1}$, and replace them by the barycenter 
$\bar p_k(T)$ computed with respect to the Fubini-Study metric on  $\P^{M-1}$.
This has the effect of reducing the purely syntactic contribution,
especially if the set of languages chosen contains languages with
sufficiently different set of syntactic parameters. Of course, it is not
possible to entirely decouple semantics from syntax, as the
syntactic-semantic interface is very rich (see for example [Ha13], [Va05]),
but this averaging method can at least partially reduce the influence of those effects
that are due to syntax alone.

\bigskip

\centerline{\bf 5. Geodesically convex neighborhoods and semantic spaces}

\medskip

In the setting above, we have associated to texts in a corpus a
collection of points (or of paths) in an ambient geometric space
(a Grassmannian, or a flag variety, or a projective space). We have
also seen that, when we group together words in the lexicon by
semantic categories, geometrically we look at how the set of
points and paths behaves under a projection map of the ambient
variety. In this section we use the same general geometric picture,
and we consider coverings by convex open sets. These local
neighborhoods correspond to grouping together texts by semantic 
similarity. The convexity property corresponds to the possibility of
interpolation and will be compared in Section 7 with the approach of G\"ardenfors
on conceptual spaces as ``meeting of minds", cf.  [G\"a00], [G\"a14], [WaG\"a13].

\smallskip

{\bf 5.1. Geodesic convexity and good coverings.} 
Recall that a subset $U\subset X$ in a Riemannian manifold $X$ 
is said to be {\it geodesically convex} if for arbitrary points $p\neq p' \in U$
there is a distance minimizing geodesic arc connecting them that is
entirely contained in $U$. In particular, a geodesically convex $U$ is
topologically a contractible set.  Moreover, a non--empty finite intersection of
geodesically convex open sets $U_i$ is also a geodesically convex
open set. If $X$ is compact, we can assume the number of open sets
in such a covering to be finite. Their size (measured as the diameter) 
in such a covering is bounded. We then say that $\cU_\epsilon=\{ U_i(\epsilon) \}_{i=1}^n$
is {\it a good $\epsilon$--covering} of the compact Riemannian manifold $X$,
if the $U_i$ are geodesically convex with 
$\epsilon=\max_i \{\roman{diam} (U_i(\epsilon)) \}$.

\smallskip

In particular, we consider such coverings for the Grassmannians
$Gr(N,M')$, flag varieties $F(1,\ldots, 1, M'-N)$, and projective spaces
$\P^{M'-1}$, with the respective metrics discussed above, and where
$M'$ is the size of the semantic vocabulary, after semantic identifications
have been performed on the initial lexical vocabulary of size $M\geq M'$,
as discussed in the previous section. 
We view points $p_k(T)$ that lie within the same convex 
neighborhood $U_i(\epsilon)$ of a good $\epsilon$-open covering 
by geodesically convex sets as being semantically related. In particular,
we are interested in considering good $\epsilon$-open coverings that are
generated by starting with a collection $U_k(\epsilon,T)$ of geodesic balls
of radius $\epsilon/2$ centered at the points $p_k(T)$ associated to a text $T$
in a corpus. Consider the case where $p_k(T)\in \P^{M'-1}$. The cases of
points in Grassmannians and flag varieties are analogous. 
We construct an $\epsilon$-open covering of the ambient variety
by starting with the collection $\{ U_k(\epsilon,T)\,:\, k=1,\ldots, N(T);\, T\in \cC \}$
and we complete it to an $\epsilon$-open covering of $\P^{M'-1}$ by adding 
enough additional sets $U_i(\epsilon)$ covering the complement of 
$Y_{\cC,\epsilon}:=\cup_{k,T} U_k(\epsilon,T)$.
We let $\cU_\epsilon$ denote the resulting covering of $\P^{M'-1}$ and we write
$\cU_\epsilon(\cC)\subset \cU_\epsilon$ for the covering of $Y_{\cC,\epsilon}\subset \P^{M'-1}$
by the $\{ U_k(\epsilon,T)\,:\, k=1,\ldots, N(T);\, T\in \cC \}$. 

\smallskip

As it is customary in topology, we can associate to a given good $\epsilon$-covering
$\cU_\epsilon(\cC)$ the simplicial complex given by its \v{C}ech complex 
$\cN_\star(\cC,\epsilon):=\cN_\star(\cU_\epsilon(\cC))$, with geometric realization
$\cN(\cC,\epsilon):=| \cN_\star(\cC,\epsilon) |$. Note that, while the geometric
realization of the \v{C}ech complex of the full covering $\cU_\epsilon$ of $\P^{M'-1}$ is
just homotopy equivalent to $\P^{M'-1}$ (see [Se68] and also [DuI] for a generalization), 
the geometric realizations $\cN(\cC,\epsilon)$ of the subcomplexes $\cU_\epsilon(\cC)$ of
$\cU_\epsilon$ in general depend on the corpus $\cC$ and will in general 
not be homotopy equivalent to the ambient space. 

\smallskip

One can then study, for a given corpus of texts $\cC$, how the homotopy type, 
and invariants such as homology, of the simplicial space $\cN(\cC,\epsilon)$ vary
with the scale $\epsilon$. According to the usual approach of persistent topology,
those features that change rapidly with the scale are attributed to random
fluctuation, while persistent features can be identified with actual structures. 

\medskip

{\bf 5.2. Geodesically convex neighborhoods, \v{C}ech complexes, and neural codes.}
As in the previous subsection, we consider simplicial complexes $\cN_\star(\cC,\epsilon)$
obtained as the \v{C}ech complex of the collection $\cU_\epsilon(\cC)$ of the 
geodesically convex balls $U_k(\epsilon,T)$ of diameter $\epsilon$ around 
the points $p_k(T)\in \P^{M'-1}$, for $k=1,\ldots, N(T)$, the number of contexts in the
text $T$ and for $T$ varying in a given corpus $\cC$. Their geometric realizations
are denoted, as above, by $\cN(\cC,\epsilon)=| \cN_\star(\cC,\epsilon) |$. 

\smallskip

Following the approach of [CuItVCYo13], we associate a code $C=C(\cC,\epsilon)$
to the collection $\cU_\epsilon(\cC)$ of geodesically convex balls. This is
a code $C\subset \{ 0,1 \}^m$, where $m=\sum_{T\in \cC} N(T)$. 
Here we assume chosen an ordering of the texts $T\in \cC$, with $n=\#\cC$,
so that we identify the set of contexts $$\{ c_1(T_1), \ldots, c_{N(T_1)}(T_1),\ldots,
c_1(T_n),\ldots, c_{N(T_n)}(T_n) \}$$ of the entire corpus $\cC$ with the set
$\{ 1, \ldots, m \}$. The code words $w \in C$ are those elements $w\in \{ 0,1\}^m$
such that 
$$ \left( \bigcap_{i\in supp(w)} U_{k_i}(\epsilon,T_i) \right) \setminus 
\left( \bigcup_{j \notin supp(w)} U_{k_j}(\epsilon,T_j) \right) \neq \emptyset, $$
where $supp(w)=\{ i \in \{ 1, \ldots, m \}\,:\, w_i=1 \}$.

\smallskip

According to the ``nerve theorem" ([Ha02], Corollary 4G.3), as discussed in 
[CuItVCYo13] and [Ma15], the homotopy type of the space 
$Y_{\cC,\epsilon}=\cup_{k,T} U_k(\epsilon,T)$ is equal to the homotopy type
of the nerve $\cN(\cC,\epsilon)$ of the complex $\cN_\star(\cC,\epsilon)$.
In particular, the persistent homology of $Y_{\cC,\epsilon}$ is the same as
the persistent homology of $\cN(\cC,\epsilon)$. 

\smallskip

This is the setting used in [CuItVCYo13] to reconstruct information about the
topology of the stimulus space from knowledge of the associated neural
code. Neural codes and the problem of how they encode the structure of
the stimulus space have been studied extensively in neuroscience, 
especially in relation to vision (see [CuItVCYo13]  and references therein). 
The study of neural codes in the linguistic setting is presently less
extensive: neural codes for syntax, based on data of neurosurgical
procedures, have been studied (see [BikSza09]). A detailed criticism of
a possible linguistic approach to neurosemantics is given for instance
in [Eli05], while a proposal for semantic representation of linguistic
data via shared neural codes (for auditory, visual or somatosensory inputs)
is analyzed in [Poe06]. 

\smallskip

We argue for a proposal of the simplicial
complexes $\cN_\star(\cC,\epsilon)$ and their persistent homotopy
type as possible computational models of neural codes for neurosemantics, 
at least up to homotopy. 
Namely, instead of the usual approach to measuring semantic relatedness of
texts on the basis of angular distances of semantic vectors, one can consider
topological notions of relatedness and proximity, in terms of deformability and
homotopy equivalence of the complexes $\cN_\star(\cC,\epsilon)$.

\bigskip

\centerline{\bf 6. Spectral decompositions and Riccati flows}

\medskip

{\bf 6.1. Singular Value Decomposition.} Typically, the word--document semantic matrices discussed above
are very sparse, with often only a small percentage of entries being
non--zero. It is known that this creates problems in measuring semantic
similarity with the usual cosine method  (see [TuPa10]), as the method
easily assigns zero to non co--occurring words even though they are
semantically related. 
\smallskip

In order to circumvent
this problem, one can perform a dimensional reduction based on a
singular value decomposition (SVD). This represents the 
semantic matrix $P$  as a product $U \Sigma V^\tau$,
where $U$ and $V$ are, respectively, and $M\times M$ and an $N\times N$ 
unitary matrix and $\Sigma$ is an $N\times M$ matrix with the singular
values on the diagonal, of rank $r$ equal to the rank of the original semantic
matrix. 

\medskip
{\bf 6.2. Latent Semantics.} The technique known as ``latent semantics" (see Section 4.3 of [TuPa10])
then considers truncations of the matrix $U \Sigma V^\tau$ to a rank $k<r$ approximation
$U_k \Sigma_k V_k^\tau$ obtained by considering only the $k$ largest singular values.
This has the effect of creating a low-dimensional linear mapping between words and contexts,
which reduces noise and improves the estimates of semantic similarity, or 
``discover latent meaning" in the terminology used in vector space semantics. 
\smallskip
Thus, according to this procedure, the process of analyzing semantic relatedness
based on the given word-context semantic matrix, involves a singular value decomposition
and a truncation according to the largest singular values. We will see in the rest of this
section that these operations also have a very natural geometric interpretation in
terms of the geometry of projective spaces and Grassmannians.

\medskip
{\bf 6.3. Term co-occurrence matrix.} In order to obtain the singular value decomposition
and restrict to the largest $k$ singular values, one considers the symmetric 
matrix $A=P^\tau P$, the term co-occurrence matrix, and its spectral decomposition. 
The truncations discussed above can then be obtained by applying power
methods to separate out the span of the eigenvectors of the largest $k$
eigenvalues of $A=P^\tau P$ from the complementary space. 
For further discussions of ``semantic spectrum" and ``eigenword" decomposition
see for instance [DhFU15], [WiDa11]. 

\medskip
{\bf 6.4. Perron--Frobenius and Riccati equation.}
If there is only one top eigenvalue one can apply the usual Perron--Frobenius theory. 
Let $Sp(A) =\{ \lambda_1, \ldots, \lambda_N \}$
with $|\lambda_1| > |\lambda_2|\geq \cdots \geq|\lambda_N|$. In the case we considered
in the previous sections where $N\leq M$ and the rank is $N$, the matrix $A$ determines
an action on $\P^{N-1}$, and the sequence of points $x_m =A^m x_0$, for an assigned
initial point $x_0\in \P^{N-1}$, converges to the point in $\P^{N-1}$ corresponding
to the line spanned by the Perron--Frobenius eigenvector of $A$. Moreover, as discussed in
[AmMa86], [MaAm92], in a local chart corresponding to vectors with first component equal 
to one, we have
$$ A: x_m =\pmatrix 1 \\ y_m \endpmatrix  \mapsto x_{m+1}=A x_m =
 \pmatrix 1 \\ y_{m+1} \endpmatrix , $$
with
$$ y_{m+1} = \frac{A_3 + A_4 y_m}{A_1 + A_2 y_m} $$
where
$$ A= \pmatrix A_1 & A_2 \\ A_3 & A_4 \endpmatrix  , $$
where $A_4$ is an $(N-1)\times (N-1)$-matrix and $A_1$ a number. The
recursion relation of the sequence $y_m$ is then given by
$$ y_{m+1} - y_m = (A_3 + A_4 y_m - y_m A_1 - y_m A_2 y_m) (A_1 + A_2 y_m)^{-1}. $$
The above can be viewed as a discretization of the matrix Riccati equation
$$ \frac{d}{dt} y(t) = A_3 + A_4 y(t) - y(t) A_1 - y(t) A_2 y(t), $$
in particular, both equations have the same stationary solutions given by solutions to 
$$ A_3 + A_4 y - y A_1 - y A_2 y =0. $$
Thus, in order to find the limit $x=\lim_m x_m$, or equivalently the stationary solution 
$y_{m+1}=y_m$ of the difference equation above, one can consider the Riccati flow
to the same fixed point. For this reformulation of the Perron--Frobenius theory in 
terms of a matrix Riccati equation in a projective space, see [AmMa86], [MaAm92].

\medskip
{\bf 6.5. Latent Semantics and flows on Grassmannians.}
In a similar way, it is shown in [AmMa86], [MaAm92] that the selection of
the span of the eigenvectors of the $k$ largest eigenvalues of the
matrix $A=P^\tau P$ can be performed dynamically in terms of a 
Riccati flow on the Grassmannian $G(k,N)$. More precisely,
for a given $k$-dimensional vector space $V\in G(k,N)$ and
a matrix $A \in GL_N$, we have $AV \in G(k,N)$ given by $AV=\{ Av\,:\, v\in V\}$.
Thus, given an initial point $V_0\in G(k,N)$ one can consider the power
sequence $V_{m+1} = A V_m$. If $Spec(A)=\{ \lambda_1, \ldots, \lambda_N \}$
with $$|\lambda_1|\geq |\lambda_2|\geq |\lambda_k| > |\lambda_{k=1}|\geq \cdots |\lambda_N|, 
$$ and $U$ is the
span of the eigenvectors corresponding to $\lambda_i$ with $i=1,\ldots, k$,
then the sequence of points $V_m$ in $G(k,N)$ converge to the point
corresponding to the space $U$, for every choice of initial $V_0$ 
with $V_0\cap W=\{ 0 \}$, where $W$ is the span of the eigenvectors with
eigenvalues $\lambda_i$ with $i=k+1,\ldots,N$. 

For a choice of complementary subspaces $U\in G(k,N)$ and $W\in G(N-k,N)$, 
and a morphism $L \in Hom(U,W)$, consider the element $U_L\in G(k,N)$ given by
the subspace 
$$ U_L =\{ \pmatrix u \\ Lu \endpmatrix \,|\, u\in U \} \subset U\oplus W. $$
If the matrix $A$ in the decomposition $U\oplus W$ has the form
$$ A = \pmatrix A_1 & A_2 \\ A_3 & A_4 \endpmatrix $$
then 
$$ A U_L = U_{(A_3+ A_4 L) (A_1 + A_2 L)^{-1}} $$
in this local chart on the Grassmannian $G(k,N)$. Thus, one obtains a
corresponding sequence
$$ L_{m+1} = (A_3+ A_4 L_m) (A_1 + A_2 L_m)^{-1} $$
which can be written as a difference equation
$$ L_{m+1} - L_m = (A_3 + A_4 L_m - L_m A_1 - L_m A_2 L_m) (A_1 + A_2 L_m )^{-1}. $$
As before, the stationary solutions can be equivalently obtained as the stationary points of the
matrix Riccati flow
$$ \frac{d}{dt} L(t) = A_3 + A_4 L(t) - L(t) A_1 - L(t) A_2 L(t). $$
This shows that the latent semantics method based on singular value
decomposition and truncation to the top $k$ singular values for $P$ 
can be reformulated in terms of a geometric flow on a Grassmannian.

\bigskip

\centerline{\bf 7.  Relation to G\"ardenfors' ``meeting of minds"}

\medskip

{\bf 7.1. Where the minds meet.} In [G\"a14], G\"ardenfors developed an approach to
semantic spaces based on the metaphor ``meeting of minds" (see [WaG\"a13])
and on models of ``conceptual spaces" developed in [G\"a00].
The main idea is that meaning is emergent in communication
(see Section 5.1 of [G\"a14]). Typically, coming to a common
understanding of meaning in communication is seen as a fixed
point problem taking place in a convex space which describes
some configuration domain, such as colors, some kind of actions,
etc. Communication is modeled in terms of a partitioning of this
domain determined by the transmitter and a sample set of points 
in the domain obtained by the received, and the common understanding
is achieved by the construction of a Voronoi partition common to
both sets of points, see Section 5.4.1 of  [G\"a14]. 

\smallskip

In our setting, the geometry of semantic spaces is not dictated by
conceptual spaces determined by preassigned external semantic
categories as in [G\"a14], but rather the geometry of  an ambient space (a Grassmannian,
or a set of paths in a projective space) built out of corpora of
texts and the frequencies of occurrences of lexemes in contexts of
these texts. However, we can still develop an approach to communication
as a fixed point problem leading to a common
semantic interpretation between different users, which resembles,
in a different geometric setting, the ``meeting of minds" approach
of G\"ardenfors.

\smallskip

Consider a set $A$ of different users. All users have access to the
same dictionary $D$ of lexemes, while each user $\alpha \in A$ has
access to a certain corpus of texts $\cC_\alpha$, and derives
semantic information from the analysis of occurrences of the
words of $D$ in the contexts of the texts $T\in \cC_\alpha$.
Thus, each user $\alpha \in A$ obtains a matrix $P_\alpha(T)$
of semantic vectors for each text $T\in \cC_\alpha$. Assuming
each user has analyzed the entire corpus $\cC_\alpha$, and
used the information available in all texts $T\in \cC_\alpha$ to
obtain semantic information, we obtain, for each lexeme $w_k \in D$
and for each user $\alpha \in A$, a semantic vector $P_{\alpha,k}=(p_{\alpha,ki})$,
where the index $i$ ranges over all the contexts $c_i(T)$ of all the
texts $T$, listed in a given order in the corpus $\cC_\alpha$. 
We can view all these semantic vectors $P_{\alpha,k}$ inside
a larger vector space that corresponds to the union $\cC =\cup_\alpha \cC_\alpha$,
where we add zero entries to the vector $P_{\alpha,k}$ whenever a certain text 
$T$ in some corpus $\cC_\beta$ is not also contained in $\cC_\alpha$. 
In this way, for a given lexeme $w_k\in D$, the different users arrive
at somewhat different semantic interpretations, depending on the
different texts they had access to. This difference is measured by
the different position of the vectors $P_{\alpha,k}$ in this ambient
space. In a similar way, if we consider the entire dictionary, or just
some subset of lexemes, we obtain for each user a different
semantic matrix $P_\alpha$, computed as above over all texts $T$
in $\cC =\cup_\alpha \cC_\alpha$. As before, we regard these matrices
as points $p_\alpha$ in a Grassmannian $Gr(M,N)$, 
where $M$ is the number of lexemes considered and $N$ is the
overall number of contexts in all the texts in the entire union $\cC$
of corpora. Here we typically are in the situation were we are
seeking a common semantic understanding of a small number 
of lexemes using a large number of context and corpora, hence $M<N$.

\smallskip

Given this finite collection $\{ p_\alpha \}_{\alpha\in A}$ of
points in a Grassmannian $Gr(M,N)$, which represents the
different positions in semantic space the different users arrived
at by analyzing the occurrence of the same list of lexemes in
the corpora available to them, we need a simple geometric procedure 
that arrives to a common position in semantic space and that
can be implemented interactively as a sequence of approximations. 
A simple such procedure consists of taking the geodesic barycenter
of the set $\{ p_\alpha \}_{\alpha\in A}$. In fact, more generally one
can considered a weighted distribution of the points $p_\alpha$,
where each $p_\alpha$ is assigned a weight $\lambda_\alpha\geq 0$
with $\sum_\alpha \lambda_\alpha =1$. The additional information
contained in the weights $\lambda_\alpha$ can be some a priori
knowledge of the higher reliability or relevance of some corpora $\cC_\alpha$
with respect to others, which would make the semantic matrix $P_\alpha$
obtained by some user more reliable than that obtained by some
other user. Given the set $\{ p_\alpha \}$ in $Gr(M,N)$ and the respective 
weights $\lambda_\alpha$, the barycenter $p_B$ is determined by the
condition
$$ \sum_\alpha \lambda_\alpha \delta^2(p_\alpha, p_B) =
\min_{p\in Gr(M,N)} \{ \sum_\alpha \lambda_\alpha \delta^2(p_\alpha, p) \}, $$
where $\delta(x,y)$ is the geodesic distance. Assuming that all
the points $p_\alpha$ lie sufficiently close to each other (as would be the case
if there is enough overlap between the corpora available to different users) 
so that they are contained in a single geodesically convex neighborhood 
$U \subset Gr(M,N)$, the potential function
$$ V(p)=\sum_\alpha \lambda_\alpha \delta^2(p_\alpha, p) $$
is a strictly convex function on the neighborhood $U$ and has therefore
a unique minimum. The barycenter is then the point $p_B$ where
$V( p)$ achieves its minimum.
\smallskip
It can be also described as the unique fixed point of the map
$p \mapsto p - h \, \nabla V(p)$,
where $\nabla V= g(dV,\cdot)$,  $g$ being the Riemannian
metric tensor, and $h$ is a finite increment in a discretized
gradient descent. Recursively, $p_B$ is then approximated by 
$p_{k+1}= p_k - h\, \nabla V(p_k)$. 

\smallskip

In a similar way, one can consider simplicial Vietoris--Rips complexes
$\cN_*(\cC_\alpha,\epsilon)$ obtained by different users based on
different corpora $\cC_\alpha$. After again considering them inside
a larger common projective space, one can construct a new complex which is
their common barycentric subdivision. The homotopy type and
persistent homology of the resulting complex can then be treated as
a model of the ``meeting of minds" in our setting.

\bigskip

\centerline{\bf 8. Semantic vectors, Zipf's law, and Kolmogorov complexity}

\medskip

{\bf 8.1. Zipf's law.} As observed in [Lowe01], constructions of semantic spaces based on
semantic vectors should take into consideration the fact that the
distribution of linguistic data is skewed towards high count data, according
to the empirical Zipf's law. 

\smallskip

Given a corpus of texts $\cC$ and a word $w$ (in the sense of a lexeme from
a dictionary of words), let $F_\cC(w)$ denote the number of tokens of the given
word that appear in the corpus, and $F_{\cC}(w)/N$ the relative frequencies,
where $N=\# \cC$ is the size of the corpus. Let $\{ w_k \}$ be an
enumeration of the dictionary words by decreasing frequencies. Then Zipf's
law states that
$\log (F_{\cC}(w_k)) = \kappa(N) - B \log(k)$, 
for a constant $\kappa(N)$ depending on the corpus size and with the
power law $B$ satisfying $B\sim 1$.  It was shown in [Ma13] that 
if one postulates that, in Zipf's original explanation as``minimization of effort'', the word
``effort'' means Kolmogorov's complexity, then Zipf's law with exponent 1
becomes a consequence of properties of 
 the related universal Levin probability distribution.

\smallskip

In the construction of semantic spaces, when one counts co--occurrences
of words in given contexts with certain given vocabulary lexemes, one
encounters a situation where low frequency words may be more significant
for semantic association, but produce very sparse semantic matrices, while
high frequency words provide more reliable statistics, but are less significant
in determining semantic association, as they tend to appear in almost every 
context. Semantic vectors based on low frequency words will have high
variance, and Zipf's law predicts that the amount of additional data
required in order to reduce the variability is expressed by a power law relation.    

\medskip

{\bf 8.2. Latent semantic analysis.} In {\it latent semantic analysis} this phenomenon is accounted for
by introducing weights assigned to the vocabulary entries, so that the
estimated probability (frequency) $p_c(w,\ell)$ of co-occurrence of a given 
word $w$ with a given lexeme $\ell$ in a context $c$ is weighted by
$S(\ell)^{-1} \log (1+p_c(w,\ell))$, 
where the denominator is given by the entropy $S(\ell)=-\sum_{c\in C(T)} p_c(\ell) \log (p_c(\ell))$,
with $p_c(\ell)$ the probability of occurrence of $\ell$ in the context $c$.
In this way, if $\ell$ is equally distributed in all contexts, as one expects for the
most frequent words, the entropy is maximal and the weighted co-occurrence
is less significant, while if $\ell$ is likely to occur only in a smaller number of contexts
the co-occurrence is weighted more, as more semantically significant. 
Other methods that can be used for taking into account the effects due to
Zipf's law and the different semantic significance of words with different frequencies
are surveyed in [Lowe01].

\smallskip

This type of considerations based on Zipf's law apply to any suitable
construction of semantic spaces, including the geometric construction
we discussed in the previous sections.  In particular, one can similarly
consider introducing appropriate weights in the construction of the
semantic matrix $P(T)$ of a text, that we discussed before, so that,
in addition to counting occurrences in contexts $c\in C(T)$, one also
keeps into account how uniform or non--uniform the distribution over
contexts is, measured in terms of the Shannon entropy of the resulting
probability distribution. 

\smallskip

\bigskip

\centerline{\bf  References}

\medskip

[AlGo95] H.~Alt, M.~Godau, {\it Computing the Fr\'echet distance between two
polygonal curves}, Int. J. Comput. Geom. Appl. Vol.5 (1995) 75--91.

\smallskip

[AmMa86] G.~Ammar, C.~Martin, {\it The geometry of matrix eigenvalue methods}, 
Acta Appl. Math. Vol.5 (1986) N.3, 239--278.

\smallskip

[Ar01] E.~Arrondo, {\it Projections of Grassmannians of lines and characterization
of Veronese varieties}, J. Alg. Geom., Vol.1 (2001) 165--192. 

\smallskip

[ArRa05] E.~Arrondo, R.~Paoletti, {\it Characterization of Veronese varieties via projections
in Grassmannians}, in ``Projective Varieties with Unexpected Properties: A Volume in Memory of Giuseppe Veronese" (C.~Ciliberto, A.V.~Geramita, B.~Harbourne, R.M.~Mir\'o-Roig, K.~Ranestad Eds.)
Walter de Gruyter, 2005.

\smallskip

[BikSza09] D.~Bickerton, E.~Szathm\'ary, {\it Biological Foundations and Origin of Syntax},
MIT Press, 2009.


\smallskip
[BuHi06] A.~Budanitsky, G.~Hirst, {\it Evaluating WordNet-based measures of semantic distance},
Computational Linguistics, Vol.32 (2006) N.1, 13--47.

\smallskip

[Ca09] G.~Carlsson. {\it Topology and data.}  Bull. AMS, vol.~46 (2009), no.~2, 255--308.

\smallskip

[ChiaBRP90] C.~Chiarello, C.~Burgess, L.~Richards, A.~Pollock, {\it Semantic and associative
priming in the cerebral hemispheres: Some words do, some words don?t $\ldots$ sometimes,
some places}, Brain and Language, Vol.38 (1990) 7--104.

\smallskip
[CuIt08] C.~Curto, V.~Itskov. {\it Cell Groups Reveal Structure of Stimulus Space.} 
PLoS Computational Biology, vol.~4, issue 10, October 2008, 13 pp. (available online).

\smallskip

[CuItVCYo13] C.~Curto, V.~Itskov, A.~Veliz-Cuba, N.~Youngs. {\it The neural ring:
An algebraic tool for analysing the intrinsic structure of neural codes.}
Bull. Math.~Biology, 75(9), pp. 1571--1611, 2013.
\smallskip

[Dem88] J.P.~Demailly, {\it Vanishing theorems for tensor powers of a positive vector bundle}, in
``Geometry and analysis on manifolds" (Katata/Kyoto, 1987), pp.86--105, Lecture Notes in Math., 
Vol.1339, Springer, 1988.

\smallskip

[DhFU15] P.~Dhillon, D.P.~Foster, L.H.~Ungar, {\it Eigenwords:
Spectral Word Embeddings}, Journal of Machine Learning Research 16 (2015)

\smallskip

[DuI] D.~Dugger, D.C.~Isaksen, {\it Hypercovers in topology}, preprint \newline
http://www.math.uiuc.edu/K-theory/0528/


\smallskip

[Eli05] C.~Eliasmith, {\it Neurosemantics and categories}, in ``Handbook of Categorization
in Cognitive Science", pp.1035--1054, Elsevier, 2005.

\smallskip

[G\"a00] P.~G\"ardenfors.  {\it Conceptual spaces: The geometry of thought.} Cambridge, Mass.
MIT Press, 2000.

\smallskip

[G\"a14] P.~G\"ardenfors.  {\it Geometry of Meaning: Semantics Based on Conceptual 
Spaces.} Cambridge, Mass. MIT Press, 2014, 343+xii pp.

\smallskip

[Gri74] P.~Griffiths, {\it On Cartan's method of Lie groups and moving frames as applied to uniqueness and existence questions in differential geometry}, Duke Math. J., Vol.41 (1974) 775--814.

\smallskip

[Gui68] P.~Guiraud. {\it The semic matrices of meaning.} Social Science Information,
7(2), 1968, pp. 131--139.
\smallskip

[Ha02] A.~Hatcher. {\it Algebraic Topology.} CUP, Cambridge, 2002.

\smallskip

[Ha13] M.~Hackl, {\it The syntax-semantics interface}, Lingua, Vol.~130 (2013) 66--87.

\smallskip

[InLe04] P.~Indefrey, W.~J.~M.~Levelt. {\it The spatial and temporal signatures of word production
components.} Cognition 92, 2004, pp.~101--144.

\smallskip

[JeLe94]  J.~D.~Lescheniak, W.~J.~M.~Levelt. {\it Word frequency effects in speech production:
retrieval of syntactic  information and of phonological form.} Journ. of Experimental Psychology: Learning,
Memory and Cognition, 20, 1994, pp.~824--843.

\smallskip

[Li] L.~Lica. {\it The Distinction between WHICH and THAT. With Diagrams.}

http://home.earthlink.net/~llica/wichthat.htm

\smallskip

[Lowe01] W.~Lowe, {\it Towards a theory of semantic space}, in ``Proceedings of the 23rd Conference of the Cognitive Science Society" (2001), pp. 576--581.   

\smallskip

[MaAm92] C.~Martin, G.~Ammar, {\it The geometry of the matrix Riccati equation and 
associated eigenvalue methods}, in ``The Riccati equation", pp.113--126, 
Comm. Control Engrg. Ser., Springer, 1991.






\smallskip

[Ma13] Yu.~I. Manin. {\it Zipf's law and L.~Levin's probability distributions.}  Functional Analysis and its Applications,
vol. 48, no. 2, 2014. DOI 10.107/s10688-014-0052-1.
Preprint arXiv:1301.0427

\smallskip

[Ma15] Yu.~I.~ Manin. {\it Neural codes and homotopy types: mathematical models of
place field recognition.}   Moscow Math. Journal, vol. 15, Oct.--Dec. 2015, pp. 1--8 . 
arXiv:1501.00897 


\smallskip

[Man12] D.~Yu.~Manin. {\it The right word in the left place: Measuring lexical
foregrounding in poetry and prose.}  www.researchgate.net

\smallskip

[MaSch99] C.D.~Manning, H.~Schuetze, {\it Foundations of statistical natural language processing}, 
MIT Press, 1999.
\smallskip

[Me16]  I.~Mel'$\roman{\check{c}}$uk. {\it Language: from Meaning to Text.}.
Ed. by D.~Beck. Moscow \& Boston, 2016.



\smallskip

[Poe06] D.~Poeppel, {\it Language: Specifying the Site of Modality-Independent Meaning},
Current Biology, Vol.16 (2006) N.21, R930--R932.

\smallskip

[PorGhGuCLDMar15] A.~Port, I.~Gheorghita, D.~Guth, J.M~Clark, C.~Liang, S.~Dasu, M.~Marcolli, 
{\it Persistent Topology of Syntax}, arXiv:1507.05134

\smallskip

[Pos06] A.~Postnikov. {\it Total positivity, Grassmannians and networks},  preprint 
arXiv:math/0609764 [math.CO].

\smallskip

[SchPe93] H.~Sch\"utze, J.~Pedersen, (1993). {\it A vector model for syntagmatic and paradigmatic
relatedness}, in ``Making Sense of Words", pp.~104--113, Oxford, 1993

\smallskip

[Se68] G.~Segal, {\it Classifying spaces and spectral sequences}, Inst. Hautes Etudes Sci. Publ. Math.
Vol.34 (1968) 105--112.


\smallskip

[TuPa10] P.D.~Turney, P.~Pantel, {\it From frequency to meaning: vector space models of semantics},
Journal of Artificial Intelligence Research, Vol.37 (2010) 141--188.

\smallskip

[Va05]  R.D.~van Valin, Jr. {\it Exploring the Syntax-Semantics Interface}, Cambridge University
Press, 2005.


\smallskip

[WaG\"a13] M.~Warglien, P.~G\"ardenfors, {\it Semantics, conceptual spaces and the meeting of minds},
Synthese, Vol.190 (2013) N.12, 2165--2193.

\smallskip

[WiDa11] P.~Wittek, S.~Dar\'anyi, {\it Spectral Composition of Semantic Spaces},
in ``Quantum Interaction",  Lecture Notes in Computer Science, Vol.7052 (2011),
pp.~60--70.

\smallskip

[Yo14] N.~E.~Youngs. {\it The neural ring: using algebraic geometry to analyse neural rings.}
arXiv:1409.2544 [q-bio.NC], 108 pp.

\vskip1cm

\enddocument